\documentclass[conference]{IEEEtran}
\IEEEoverridecommandlockouts
\usepackage{cite}
\usepackage{amsmath,amssymb,amsfonts}
\usepackage{algorithmic}
\usepackage{graphicx}
\usepackage{textcomp}
\usepackage{booktabs}
\usepackage{xcolor}

\def\BibTeX{{\rm B\kern-.05em{\sc i\kern-.025em b}\kern-.08em
    T\kern-.1667em\lower.7ex\hbox{E}\kern-.125emX}}
\begin{document}

\title{LeJOT: An Intelligent Job Cost Orchestration Solution for Databricks Platform}

\title{LeJOT: An Intelligent Job Cost Orchestration Solution for Databricks Platform}

\author{\IEEEauthorblockN{Lizhi Ma, Yi-Xiang Hu, Yuke Wang, Yifang Zhao, Yihui Ren, Jian-Xiang Liao, Feng Wu\IEEEauthorrefmark{1}, Xiang-Yang Li\IEEEauthorrefmark{1}\thanks{\IEEEauthorrefmark{1}Corresponding authors.}}
\IEEEauthorblockA{
\textit{University of Science and Technology of China}, Hefei, China\\
\{malizhi,yixianghu,yk\_wang,zhaoyifang\}@mail.ustc.edu.cn, \\\{renyh10,liaojx3\}@lenovo.com, \{wufeng02,xiangyangli\}@ustc.edu.cn}}
\maketitle

\begin{abstract}

With the rapid advancements in big data technologies, the Databricks platform has become a cornerstone for enterprises and research institutions, offering high computational efficiency and a robust ecosystem. However, managing the escalating operational costs associated with job execution remains a critical challenge. Existing solutions rely on static configurations or reactive adjustments, which fail to adapt to the dynamic nature of workloads. To address this, we introduce LeJOT, an intelligent job cost orchestration framework that leverages machine learning for execution time prediction and a solver-based optimization model for real-time resource allocation. Unlike conventional scheduling techniques, LeJOT proactively predicts workload demands, dynamically allocates computing resources, and minimizes costs while ensuring performance requirements are met. Experimental results on real-world Databricks workloads demonstrate that LeJOT achieves an average 20\% reduction in cloud computing costs within a minute-level scheduling timeframe, outperforming traditional static allocation strategies. Our approach provides a scalable and adaptive solution for cost-efficient job scheduling in Data Lakehouse environments.
\end{abstract}

\begin{IEEEkeywords}
cost optimization, performance prediction, solver, cloud computing
\end{IEEEkeywords}

\section{Introduction}

The Data Lakehouse~\cite{armbrust2021lakehouse} architecture has emerged as a transformative paradigm in data management, seamlessly integrating the scalability and flexibility of data lakes with the structured performance and governance benefits traditionally found in data warehouses. This hybrid model effectively resolves issues associated with data lakes—such as inconsistent data quality and weak governance—while preserving cost efficiencies through the separation of storage and compute resources~\cite{databricksBestPractices}. As organizations increasingly leverage the Data Lakehouse architecture to support advanced analytics, machine learning (ML), and real-time application workloads, the optimization of operational costs has become essential for maintaining competitiveness and sustainable growth.

Operational cost management in Data Lakehouse environments involves a complex interplay of factors~\cite{key}, including data storage management, efficient utilization of computing resources, workload optimization, and adaptive query management~\cite{katari2021cost}. Databricks, as a prominent Data Lakehouse platform, provides several best practices for cost control~\cite{databricksBestPractices}: selecting appropriate resources based on job characteristics, dynamically scaling resources in response to changing demands, continuously monitoring resource utilization for inefficiencies, and designing workflows~\cite{s21248212} that emphasize cost-effective query execution. However, despite clear guidelines, organizations often struggle to practically implement these strategies due to the highly dynamic and unpredictable nature of modern data workloads.

A critical challenge lies in balancing resource provisioning and performance. Organizations frequently resort to over-provisioning to guarantee performance during peak usage periods, which inevitably results in resource waste and increased operational costs. Conversely, under-provisioning resources risks compromising application performance and user experience. Additionally, manual resource management processes are labor-intensive, prone to human error, and unsustainable as workloads scale in complexity and diversity. Therefore, there is a pressing need for intelligent, automated approaches capable of dynamically predicting and allocating resources to achieve cost efficiency without sacrificing performance.

In response to these challenges, we propose LeJOT, an intelligent, solver-based job orchestration solution specifically designed for optimizing resource allocation and minimizing costs within Databricks environments. Unlike traditional scheduling techniques that depend heavily on static configurations or reactive adjustments, LeJOT leverages predictive analytics combined with real-time optimization algorithms to proactively allocate resources. By predicting job execution times through lightweight ML models and employing sophisticated solver-based optimization methods, LeJOT dynamically manages resources, ensuring that allocations precisely match workload requirements. This proactive approach not only reduces operational costs but also guarantees that individual job performance constraints are consistently satisfied.

Our contributions in this paper are summarized as follows:
\begin{itemize}
    \item \textbf{Cost Optimization Formulation}: We propose the first comprehensive mathematical model explicitly targeting cost optimization in Data Lakehouse environments.
    \item \textbf{Execution Time Prediction}: We introduce a lightweight yet highly accurate ML model to predict job execution durations across various resource configurations.
    \item \textbf{Solver-based Scheduling Framework (LeJOT)}: We present an intelligent solver-based orchestration framework that dynamically optimizes job execution costs in real-time.
    \item \textbf{Experimental Validation}: We rigorously evaluate LeJOT against real-world Databricks workloads, demonstrating a significant cost reduction averaging over 20\% compared to traditional static and reactive strategies.
\end{itemize}


\section{Background and Motivation}

With the rapid expansion of big data analytics, the demand for scalable, cost-effective data processing solutions has significantly increased. Databricks~\cite{databricksDatabricksDocumentation}, a leading unified data analytics platform based on Apache Spark, is widely adopted for its robust capabilities in building, deploying, and maintaining enterprise-grade data, analytics, and artificial intelligence applications~\cite{ilijason2020beginning}. Despite its extensive features, Databricks faces a persistent mismatch between allocated computational resources and actual user job execution needs, leading to resource underutilization and consequently escalating operational costs~\cite{492493}.

A primary reason for such inefficiencies is the lack of precise correlation between allocated computational resources and the anticipated execution times of tasks~\cite{abid2020challenges}. Users often default to standard resource configurations provided by Databricks, failing to tailor allocations to the specific complexities and demands of their workloads. Although Databricks includes dynamic resource scaling capabilities, these are generally reactive rather than proactive, making optimal resource usage and cost efficiency elusive in practical scenarios.

This resource allocation challenge becomes more acute given the variability in data processing workloads, characterized by fluctuating intensity and diversity~\cite{bal2022joint,eleliemy2021resourcefulcoordinationapproachmultilevel}. Traditional methods typically involve static resource provisioning or reactive adjustments, both of which have inherent limitations. Static approaches frequently lead to resource wastage through over-provisioning to accommodate peak demands, while reactive methods often fall short in addressing rapid workload changes, negatively impacting application performance and user experience. Additionally, manual tuning of resources is labor-intensive, error-prone, and unsustainable at scale.

Although existing literature provides theoretical insights into resource management and task scheduling strategies—such as dynamic allocation models based on historical predictions~\cite{DBLP:journals/corr/abs-2109-09269}, heuristic scheduling algorithms~\cite{spitzen2019job}, and ML applications~\cite{9492638,WAUTERS201492} to enhance resource utilization—these methodologies remain predominantly theoretical, with limited practical solutions tailored explicitly to real-world Databricks environments.

In response to these challenges, we introduce LeJOT, an intelligent, solver-based orchestration framework specifically designed for Databricks platforms. LeJOT integrates predictive analytics through ML to accurately estimate job execution times and leverages sophisticated optimization algorithms to dynamically allocate resources. This approach ensures high resource utilization, maintains performance criteria, and substantially reduces operational costs. Through a proactive, predictive strategy, LeJOT addresses the critical need for automated, efficient resource management solutions, filling the existing gap between theoretical research and practical application in Databricks environments.

\section{Formulation}

In this section, we formally describe the optimization model studied in this paper. We consider a set of workflows \( W \), where each workflow \( i \) must start no earlier than its earliest start time \( e_i \) and finish no later than its deadline \( l_i \). Additionally, there is a set of device types \( D \), with each device type \( d \) having a set of available configurations \( K_d \). For each device type \( d \) and configuration \( k \), the estimated number of devices required is denoted by \( b_{d,k} \). The expected execution time \( h_{i,d,k} \) for each workflow \( i \) varies depending on the device type \( d \) and configuration \( k \) (see Section \ref{exe}). Moreover, the price of each device \( d \) depends on its usage: when the usage is below the pre-purchased threshold, the unit price is \( c_d^0 \); however, once the usage exceeds the threshold, the unit price increases to \( c_d^1 \), following a tiered pricing structure.

The goal of this problem is to determine the start time \( s_i \) and select a device configuration \( x_{i,d,k} \) for each workflow \( i \in W \) such that the total resource costs are minimized, while meeting the constraints outlined below.

The symbols and corresponding definitions used in this paper are listed in Table \ref{tab2}. The mathematical formulation of the model is as follows:


\begin{equation}
    \min \sum_{d \in D} (u_d^1 - A_d) \times (c_d^1 - c_d^0) + \sum_{d \in D} u_d \times c_d^0
\end{equation}
subject to the following constraints:
\begin{equation}
    \sum_{d,k} x_{i,d,k} = 1, \quad \forall i \in W
\end{equation}
\begin{equation}
    \sum_{i,k} h_{i,d,k} \times b_{d,k} \times x_{i,d,k} = u_d, \quad \forall d \in D
\end{equation}
\begin{equation}
    u_d \leq u_d^1, \quad \forall d \in D
\end{equation}
\begin{equation}
    A_d \leq u_d^1, \quad \forall d \in D
\end{equation}
\begin{equation}
    s_i + g_i = t_i, \quad \forall i\in W
\end{equation}
\begin{equation}
    \sum_{d,k} h_{i,d,k} \times x_{i,d,k} = g_i, \quad \forall i\in W
\end{equation}
\begin{equation}
    t_i  \leq s_j, \quad \forall (i,j) \in P
\end{equation}
\begin{equation}
    e_i \leq s_i, \quad \forall i \in W
\end{equation}
\begin{equation}
    t_i \leq l_i, \quad \forall i \in W
\end{equation}


    \textbf{Device Configuration Constraint (Equation 2).} For each workflow \( i \in W \), exactly one device configuration \( d \) and \( k \) must be selected. This ensures that each workflow is assigned to a specific device and configuration.

    \textbf{Device Usage Constraint (Equation 3).} The total usage \( u_d \) of each device \( d \in D \) is computed by summing over all workflows \( i \in W \) and configurations \( k \). The usage is weighted by the expected execution time \( h_{i,d,k} \) and the estimated number of devices \( b_{d,k} \), ensuring that the total usage for each device is correctly calculated.

    \textbf{Device Capacity Constraint (Equation 4).} The total usage \( u_d \) for each device type \( d \in D \) must not exceed the maximum allowable usage \( u_d^1 \), ensuring that devices are not overused beyond their capacity.

    \textbf{Pre-purchased Device Usage Constraint (Equation 5).} The number of devices pre-purchased for each device type \( d \), denoted as \( A_d \), must be less than or equal to the maximum allowed usage \( u_d^1 \). This constraint ensures that pre-purchased devices are properly accounted for in the total usage calculation.

    \textbf{Workflow Start and Finish Time Constraint (Equation 6).} This constraint links the start time \( s_i \) and finish time \( t_i \) of each workflow \( i \in W \). The finish time is calculated as the start time plus the execution time \( g_i \), ensuring that each workflow’s duration is correctly accounted for.

    \textbf{Workflow Execution Time Constraint (Equation 7).} The total execution time \( g_i \) for each workflow \( i \) must match the sum of execution times across all selected device configurations, ensuring that the execution time is consistent with the resources allocated.

    \textbf{Precedence Constraint (Equation 8).} This constraint ensures that if workflow \( i \) must be completed before workflow \( j \) (denoted by the pair \( (i,j) \in P \)). This enforces the required precedence relationships between workflows.

    \textbf{Earliest Start Time Constraint (Equation 9).} Each workflow \( i \in W \) must start no earlier than its earliest allowed start time \( e_i \), ensuring that workflows are not scheduled too early.

    \textbf{Deadline Constraint (Equation 10).} Each workflow \( i \in W \) must finish no later than its deadline \( l_i \), ensuring that all workflows complete within their specified deadlines.

Together, these constraints define the feasible solutions to the optimization problem, ensuring that workflows are assigned devices and configurations that respect all constraints, while minimizing the total cost of using the devices.


\begin{table*}[tb]
\caption{Symbols and Definitions}
\begin{center}
\begin{tabular}{l|l}
\toprule
\textbf{Symbol} & \textbf{Definition} \\
\midrule
$W$  & Set of workflows, where each workflow represents a task to be completed. \\
$P$  & Set of precedence constraints between workflows, denoted as pairs \( (i,j) \). \\
$D$  & Set of available device types, each corresponding to a different kind of hardware. \\
$s_i$  & Start time of workflow \( i \), indicating when the workflow begins execution. \\
$t_i$  & Finish time of workflow \( i \), indicating when the workflow completes execution. \\
$e_i$  & Earliest possible start time for workflow \( i \), ensuring it starts no earlier than this time. \\
$l_i$  & Latest possible finish time for workflow \( i \), ensuring it finishes no later than this time. \\
$c_d^0$  & Base cost per hour of using device \( d \), applied when the device's usage is below the pre-purchased threshold. \\
$c_d^1$  & Incremental cost per hour of using device \( d \), applied when the device's usage exceeds the pre-purchased threshold. \\
$u_d$ & Total usage of device \( d \), representing the total time the device is used across all workflows. \\
$u_d^1$ & Maximum allowed usage for device \( d \), above which the device incurs additional costs. \\
$A_d$ & Number of devices of type \( d \) pre-purchased, impacting the total cost calculation. \\
$K_d$ & Set of available configurations for device \( d \), representing different ways the device can be set up for workflow execution. \\
$x_{i,d,k}$ & Binary decision variable: \( x_{i,d,k} = 1 \) if workflow \( i \) is assigned to device \( d \) with configuration \( k \), and \( x_{i,d,k} = 0 \) otherwise. \\
$h_{i,d,k}$ & Execution time for workflow \( i \) on device \( d \) using configuration \( k \). \\
$g_i$ & Total execution time of workflow \( i \), computed as the sum of execution times across all selected device configurations. \\
$b_{d,k}$ & Estimated number of devices required for configuration \( k \) of device \( d \). This reflects the device's expected usage. \\
\bottomrule
\end{tabular}
\label{tab2}
\end{center}
\end{table*}


\section{LeJOT Framework}
LeJOT aims to estimate the execution time of
scheduling jobs based on historical data and job characteristics. By using the estimated time and user-defined constraints, the method schedules tasks while meeting both
user requirements and general operational constraints, ensuring full resource utilization. The goal is to recommend
a reasonable resource allocation for job execution, achieving the lowest possible execution cost. To realize this cost
optimization objective, two main tasks need to be accomplished: Estimate Execution Time: Determine the predicted execution time of the task under different resource allocations. Optimize Resource Allocation: Within the constraint of meeting the estimated time requirements, output
the resource allocation that results in the lowest cost.

\begin{figure*}[htbp]
\centerline{\includegraphics[width=0.95\textwidth]{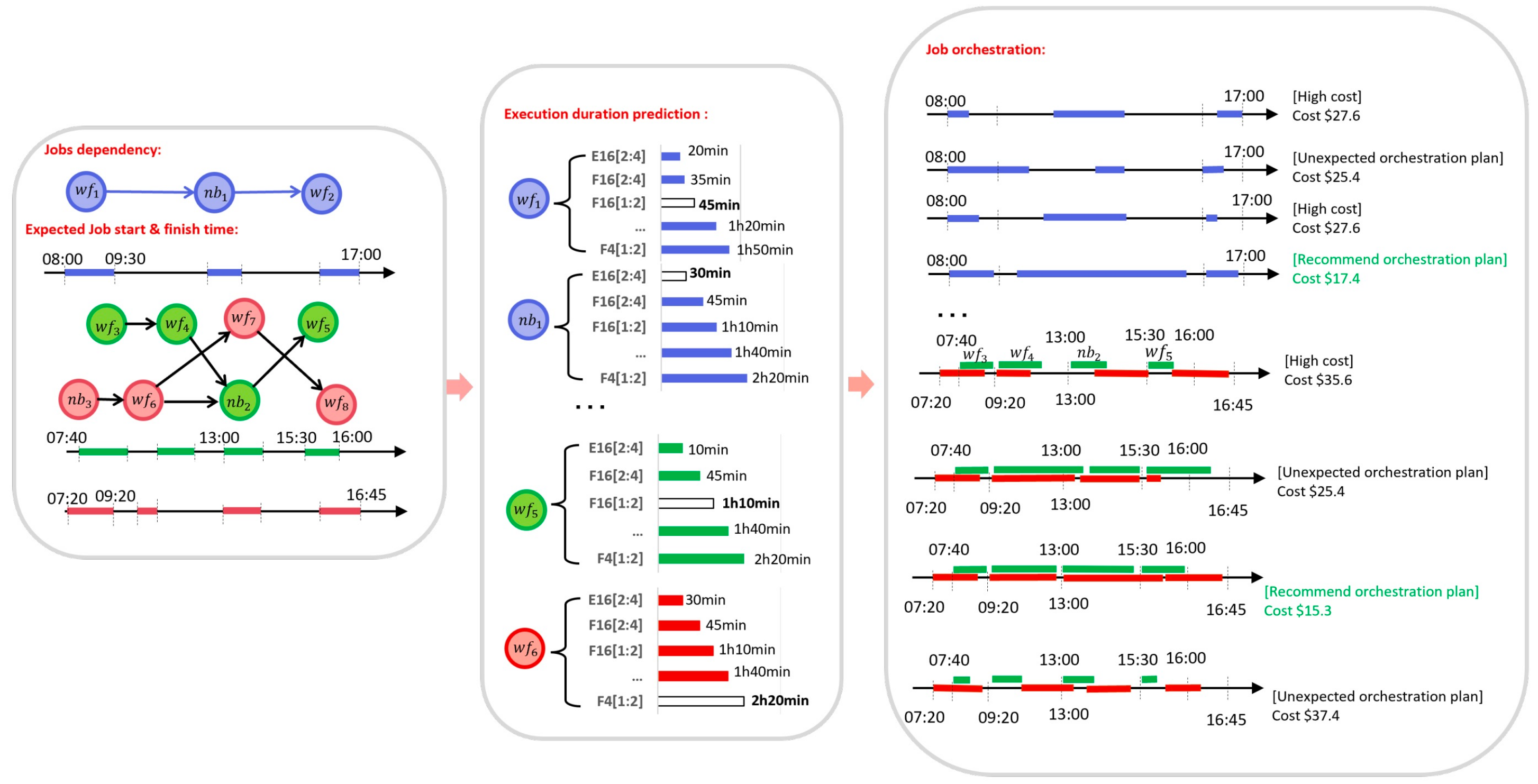}}
\caption{The overview of LeJOT framework. The diagram is divided into three sequential
execution parts: (a) Input: Provide the earliest start time and latest end time for job expectations, as well as their dependencies between
jobs. (b) Runtime Prediction: Use ML algorithms to predict each job’s runtime under different resource allocations. (c) Optimal Resource Allocation: Solve for the lowest cost resource allocation that meets dependencies and time constraints.}
\label{fig}
\end{figure*}

As shown in Figure \ref{fig}, in the input phase, users need to
provide the earliest start time and latest end time for job expectations, as well as explicitly define the dependencies
between jobs. These inputs serve as foundational data for
subsequent prediction and optimization processes.
During the ML prediction stage, historical
data is collected and utilized to model and predict the runtime of each job under different resource conditions using
ML algorithms. Feature engineering plays a
critical role in this process, involving the extraction of relevant information such as job complexity and dependency
relationships to enhance the accuracy of predictions.
In the optimization solving phase, based on estimated
execution times and user time constraints, a cost-optimal
solver is employed to determine the minimum cost resource
allocation that satisfies all time restrictions. This process
takes into account the feasibility and efficiency of resource
distribution to ensure optimal configuration under given constraints.
Through the collaborative efforts of these three phases,
the system effectively recommends resource allocation
schemes that not only meet job time requirements but also
minimize execution costs.

LeJOT primarily consists of two parts: (1)
data processing and training related to the execution time
prediction model; (2) constraint definitions and optimization objectives for the job orchestration algorithm. We will describe these two parts in detail below.

\subsection{Job Execution Duration Time Prediction Model}\label{exe}
The execution time prediction module is designed to estimate the duration of job execution using ML
model trained on historical data.

\subsubsection{Factors that Impact the Duration}
 The model receives input features including: Resource Type: Number of CPU cores and memory
size. Job Parallelism: Degree of parallel processing for the
job. Subtask Count: Number of subtasks within the job. Table Count: Number of data tables involved in the logic processing.
Code Length: Size of the code executed by the job. Job
Code: Specific code used in the job. Involved Datasets
: Data volume of tables used in the job. Data Volume on
Disk: Actual size of data tables on disk. Task Type: I/O-intensive or computation-intensive type of task. Historical
Execution Logs: Execution duration of the same job under
different resource allocations (stored in DB). These features
are carefully selected to capture key aspects influencing job
execution time.

\subsubsection{Data Collection}
Data is collected from two main
sources: 
\begin{itemize}
    \item \textbf{API Interface}: A portion of the data is retrieved
using Databricks’ public API interface. This includes information such as: Resource allocation details for workflows
and tasks (e.g., CPU and memory specifications). Number of
tasks in a workflow. Code executed by each task.
\item \textbf{Manual
Collection}: Another portion of data is collected manually,
including: Data tables used in jobs and their sizes on disk.
Task type classification (I/O or computation-intensive). 

\end{itemize}

All data originates from Lenovo’s operational experience with the Databricks platform, ensuring practical relevance.
\subsubsection{Correlation Analysis}
Through the analysis of the correlation between design features and prediction targets, this
study identifies that the three most highly correlated features are: Task Count: The number of subtasks in a job
directly impacts execution time due to parallel processing
overhead. Resource Allocation: CPU cores and memory size significantly affect task completion speed. Data Volume on Disk: Larger datasets require more time for I/O operations, especially in distributed computing environments.

\subsubsection{Model Training}

\begin{itemize}
\item \textbf{Model Select}: This paper addresses the performance optimization requirements for Lenovo's job scheduling scenarios by systematically constructing a multi-dimensional regression analysis framework. During the model selection phase, seven regression models, including linear regression, Lasso regression, and elastic net, were thoroughly evaluated. Ultimately, Ridge Regression was selected as the core algorithm based on the following critical factors:
Firstly, the feature space of the job scheduling scenario exhibits significant high-dimensionality characteristics. Traditional OLS (Ordinary Least Squares) regression can lead to severe overfitting when encountering an irreducible $X^{\mathrm{T}}X$ matrix. In contrast, Ridge Regression introduces an L2 regularization term, which imposes convex optimization constraints in the loss function. This not only ensures the uniqueness of the solution but also effectively mitigates the "curse of dimensionality."
Secondly, industrial scenarios demand high robustness against outliers. Through Monte Carlo simulations, we observed that when production environment log data contains 5\% outliers, the Mean Absolute Error (MAE) metric for Lasso Regression fluctuates by 18.7\%, whereas Ridge Regression only experiences a 6.2\% variation. This superior performance is attributed to the smooth adjustment of feature weights via its parameter shrinkage mechanism.
Lastly, in terms of inference efficiency, benchmark tests using SparkML demonstrate that Ridge Regression achieves a significant reduction in inference latency on an 8-core CPU, meeting the second-level response requirements for real-time scheduling in production systems.

 \item \textbf{Hyperparameter}: In terms of hyperparameter optimization, we designed a hierarchical grid search strategy for tuning the regularization coefficient ${\alpha}$: First, a coarse-grained scan is conducted in the logarithmic space from $10^{-2}$ to $10^{1}$ with a step size of 101. Once the optimal interval is identified, the search switches to a linear fine-grained search with a step size of 0.01. To avoid falling into local optima, we innovatively introduced a Bayesian optimization module, which uses a Gaussian process surrogate model to establish a probabilistic mapping between ${\alpha}$ values and cross-validation scores. After 200 iterations, the algorithm converges to the global optimal ${\alpha}$ = 1.45.

 \item \textbf{Training Strategy}: During the training process, cross-validation and an ensemble of multiple models are employed to mitigate the impact of random data splits on model stability. Additionally, during the splitting phase, efforts are made to ensure that each type of machine configuration maintains balanced proportions within the training set.
\end{itemize}

\subsection{Job Scheduling Optimization Algorithm}
For the job scheduling module, we have already detailed the formulation of the optimization objective (Equation 1), and all constraints (Equation 2 $\sim$ Equation 10). The specific solution method used is the Integer programming Branch and Bound, B\&B method. This approach is suitable for optimization problems where some or all decision variables are restricted to integer values. In this algorithm, certain recommended configurations, such as the number of computing resources (workers), must be integers, making it highly applicable for solving with this method.
The core idea of this approach is to decompose the main problem into multiple subproblems, solve each subproblem individually, and then integrate their solutions to obtain the solution to the main problem. Specifically, the branch-and-bound (B\&B) method involves the following key steps:
\begin{itemize}
 \item \textbf{Initialization}: Start with the original problem by relaxing the integer constraints, converting it into a linear programming (LP) problem for initial solving. At this stage, decision variables can take non-integer values.
 \item \textbf{Feasibility Check}: If the optimal solution of the LP satisfies all integer constraints, it is directly considered as the optimal solution to the original problem; otherwise, proceed to the branching step.
 \item \textbf{Branching}:  Select a decision variable that does not satisfy the integer constraint and split it into two new subproblems by restricting it to be less than or equal to its current value (floor) or greater than or equal to its current value plus one (ceiling). Each subproblem corresponds to a new branch, which is then solved separately.
 \item \textbf{Recursive Solving}:  Repeat steps 1-3 for each subproblem until all subproblems have optimal solutions that satisfy the integer constraints or no better feasible solutions exist.
 \item \textbf{Bounding (Pruning)}: During the solving process, if the optimal value of a subproblem is already greater than or equal to the current known optimal value, this branch can be discarded to avoid unnecessary computations. This strategy helps improve algorithm efficiency and reduce the search space.
 \item \textbf{Integrating Solutions}: Compare all feasible solutions that satisfy the integer constraints and select the best one as the optimal solution to the original problem.
 \end{itemize}

The advantages of the B\&B method lie in its systematic exploration of the entire solution space and its ability to efficiently reduce computational effort through pruning strategies. This method performs exceptionally well in solving problems with discrete decision variables, particularly those involving resource allocation or task assignment. In our job scheduling module, due to the involvement of multiple integer decision variables, the B\&B method can effectively find optimal configurations that satisfy all constraints.

By decomposing the main problem into subproblems and combining systematic pruning strategies, the B\&B method ensures efficiency and accuracy when solving complex optimization problems. This approach is not only capable of handling small-scale problems but also demonstrates good applicability for medium-sized and even large-scale integer programming problems.
The scheduling algorithm selects resources based on the execution times of each unscheduled task under different resource conditions and schedules tasks accordingly. Subject to the constraints of task dependencies and time limitations, the scheduling algorithm seeks to achieve an optimal cost
objective.

\section{Experiments}
In order to validate the effectiveness of the job scheduling algorithm in terms of its scheduling outputs, such as changes in job resource cost, this study adopts a quantitative research approach and a positivist research paradigm. Numerical metrics are systematically analyzed to objectively assess the algorithm's performance and validate its practical impact.
The scheduling algorithm’s experimental dataset is derived from Lenovo’s historical data on the Databricks platform, spanning from November 2024 to January 2025. On average, there are approximately 50,000 runtime records per
month. To clearly observe the optimization process and
evaluate the optimization effects of the algorithm, this paper processed the data by removing duplicates and cleaning
it. Additionally, the cost comparison before and after scheduling was analyzed.

\subsection{Evaluation Metrics}
In terms of evaluation metrics, it is mainly divided into two parts. One part is the evaluation metrics for time prediction models, and for this model, currently using common metrics in ML regression models, eg MAE, MSE (Mean Squared Error) etc. The other part is the evaluation metrics for job scheduling algorithms, and for these metrics, the paper will assess based on execution efficiency, effectiveness of scheduling results, as well as scheduling failure rate.

For the job execution duration prediction model, this paper employs common ML regression algorithm metrics, including MAE, MSE, MAPE (Mean Absolute Percentage Error), and RMSE (Root Mean Squared Error), to evaluate model accuracy. Additionally, various regression models such as linear regression, LASSO regression, elastic net, and ridge regression are validated for their prediction accuracy in this scenario. The experiment uses Expert EXP (Experience) as the baseline. On this basis, evaluate the performance of each model and optimize it. Not only is the model accuracy assessed, but it also conducts tests on inference efficiency. Specifically, it evaluates inference speed using SparkML across varying data volumes (1k,5k,10k) and time-based prediction models, while maintaining identical resource configurations (CPU: 8 cores, Memory: 32GB).

MAE measures the average absolute difference between predicted values and true values.
\begin{equation}\label{eq1266}
\mathrm{MAE}=\frac{1}{n} \sum_{i=1}^{n}\left|y_{i}-\hat{y}_{i}\right|
\end{equation}
where $y_{i}$ is the true execution time (in seconds) of sample $i$,
$\hat{y}_{i}$ is the predicted execution time (in seconds) of sample $i$,
$n$ is the total number of samples.

\begin{equation}\label{eq111}
\mathrm{MSE}=\frac{1}{n} \sum_{i=1}^{n}\left(y_{i}-\hat{y}_{i}\right)^{2}    
\end{equation}
MSE measures the average squared difference between predicted execution times and true execution times, in seconds squared.

\begin{equation}\label{eq1277}
\mathrm{MAPE}=\frac{1}{n} \sum_{i=1}^{n}\left|\frac{y_{i}-\hat{y}_{i}}{y_{i}}\right| \times 100 \%
\end{equation}
MAPE measures the average percentage error between predicted execution times and true execution times, expressed as a percentage.

\begin{equation}\label{eq1288}
\mathrm{RMSE}=\sqrt{\frac{1}{n} \sum_{i=1}^{n}\left(y_{i}-\hat{y}_{i}\right)^{2}}
\end{equation}
RMSE is the square root of MSE and measures the average difference between predicted execution times and true execution times, in seconds. 

\begin{table*}[tb]
\caption{Job Execution Duration Time Prediction Model Performance and Computing Time on Different Scales of Instances ($n$)}
\begin{center}
\begin{tabular}{l|c|c|c|c|c|c|c}
\toprule
\textbf{Model}&  \textbf{MAE}&\textbf{MSE} &\textbf{MAPE} &\textbf{RMSE} &\textbf{$n=1000$ (s)} &\textbf{$n=5000$ (s)} &\textbf{$n=10000$ (s)}\\
\midrule
Expert EXP (Baseline) & 452.09 & 705208.87 & 23.77\% & 839.76 & -   & -&-\\
Linear Regression & 272.72 & 219110.41 & 20.34\% & 468.09 & 0.3& 1.2&2.3\\
LASSO Regression & 268.41 & 215643.78 & 19.82\% & 464.35 & 0.4& 1.5&2.8\\
Support Vector Regression & 245.67 & 198754.32 & 18.15\% & 445.83 & 5.8& 28.7&57.4\\
Random Forest Regression & 212.54 & 142872.47 & 15.76\% & 412.32 & 3.2& 16.1&32.5\\
Ridge Regression & 223.72 & 182721.32 & 17.24\% & 428.62 & 0.5& 1.8&3.4\\
Elastic Net & 258.37 & 208954.21 & 19.45\% & 457.11 & 0.6& 2.1&4.0\\
Gradient Boosting Decision Tree & 203.69 & 125432.19 & 15.05\% & 401.29 & 8.5& 42.3&84.7\\
\bottomrule
\end{tabular}
\label{tab3}
\end{center}
\end{table*}

For scheduling algorithms, assessing the performance and accuracy of scheduling algorithms is crucial, as it provides valuable feedback during the constraint and optimization processes. In terms of
scheduling performance, this paper evaluates the algorithm
based on Metric Throughput, as $T$ shown in Equation \ref{eq11}
means the number of jobs processed per second, measured
in jobs/second. $N$ is the total number of jobs successfully
completed during a specific time period. $t$ is the total time
taken to process all the jobs, measured in seconds.
\begin{equation}\label{eq11}
    T=\frac{N}{t}
\end{equation}

Reliability: This refers to the algorithm’s ability to correctly execute tasks, including its performance in handling failures and abnormal conditions.
\begin{equation}\label{eq122}
    R=\frac{S+F}{T}
\end{equation}

As shown in Equation \ref{eq122}, $R$ denotes the system’s reliability, $S$ indicates the count of successful tasks, $F$ refers to the
number of tasks successfully restored via the fault-tolerant mechanism, and $T$ signifies the overall number of scheduled job requests.
In addition, cost is one of the primary optimization objectives for this paper, measuring the changes in cost by calculating the savings before and after scheduling through a cost variability rate.

\begin{equation}\label{eq13}
    CCR=\frac{FC-IC}{IC}
\end{equation}
In Equation \ref{eq13}, $CCR$ stands for the Cost Change Rate, representing the proportion of reduced costs achieved through scheduling algorithms relative to the unscheduled costs, FC represents the final cost, and IC stands for the initial cost, which refers to the calculated costs before scheduling algorithms are applied.

\subsection{Computing Environment}
The entire development process of this paper is divided
into two parts: the development of the time prediction
model and the development of the job scheduling optimization algorithm. Specifically, the experimental environment
for the job scheduling optimization algorithm is in Table \ref{tab:pre}. Similarly, the experimental environment for the time prediction model is also described in \ref{tab:my_label}.
\begin{table}[tb]
    \centering
    \caption{Experimental environment for Job Orchestration}
    \label{tab:pre}
    \begin{tabular}{l|l}
    \toprule
         Item & Detail\\
         \midrule
         CPU &AMD EPYC 7763 64-Core\\
Memory& 512G\\
Operation System& Ubuntu 22.04.3 LTS\\
         \bottomrule
    \end{tabular}
\end{table}
\begin{table}[tb]
    \centering
    \caption{Experimental environment for Time Prediction Model}
    \label{tab:my_label}
    \begin{tabular}{l|l}
    \toprule
         Item & Detail\\
         \midrule
         CPU &i5-13500H 8-Core\\
Memory& 32G\\
Operation System& Windows 11 Professional\\
         \bottomrule
    \end{tabular}
\end{table}
 To solve cost optimization problems in LeJOT, we use SCIP 8.1.0~\cite{bestuzheva2021scipoptimizationsuite80}.

\subsection{Evaluation}
This paper utilizes data from Databricks spanning from November 2024 to January 2025 for testing purposes.

For time prediction models, the Gradient Boosting Decision Tree has the highest accuracy but performs relatively slowly in terms of inference speed, failing to meet the industrial requirement for second-level response in real-world scenarios. Besides, models like Linear Regression or LASSO Regression have very fast inference speeds but are overly simplistic and perform poorly in terms of prediction accuracy. Therefore, considering both factors comprehensively, Ridge Regression was ultimately selected as the primary model for the current time prediction module. It demonstrates relatively excellent performance across various metrics and inference speed.

For scheduling optimization algorithms, the testing encompasses three key indicators: throughput to assess the algorithm’s calculation speed, reliability to evaluate its robustness, and CCR to measure cost optimization changes. Furthermore, specific pre-algorithm and post algorithm costs are analyzed.
As shown in Table \ref{tab4}, it can be observed that this algorithm possesses an extremely fast scheduling speed and a
very high scheduling success rate. The job scheduling success rate is generally above 97\%, which benefits from the
good accuracy of the execution time prediction model. Additionally, the monthly execution cost reduction is universally over 20\%, which will save most computing costs for
enterprises.

\begin{table}[tb]
\caption{Experimental results}
\begin{center}
\begin{tabular}{l|l|l|l|l|l|l}
\toprule
\textbf{Cases}&  \textbf{Datasize}&\textbf{T (k/s)} &\textbf{R (\%)}&\textbf{IC (\text{¥})}&\textbf{FC (\text{¥})}&\textbf{CCR}\\
\midrule
 2024.11& 7.05k& 1.17& 100&52.6k& 42.6k& 0.19\\
 2024.12& 57.8k& 0.103& 100&446.7k& 352.9k& 0.21\\
 2025.01& 58.84k& 0.11& 100& 426.9k& 350.1k& 0.18\\
\bottomrule
\end{tabular}
\label{tab4}
\end{center}
\end{table}


The significant performance gains can be attributed to the accurate predictive capabilities of the job execution time model, facilitating precise and proactive resource allocation. Throughput values illustrate that, despite the considerable data size increase from November to January, LeJOT maintains stable and efficient performance. However, the slight reduction in throughput with increased workload highlights areas where further algorithmic refinement could enhance scalability.
The achieved cost reduction consistently around 20\% emphasizes the framework’s practical applicability and potential to substantially decrease operational costs for enterprises operating in Data Lakehouse environments. Future improvements could focus on further enhancing prediction accuracy for highly complex tasks and extending the optimization model to handle GPU-intensive workloads and multi-cloud scenarios, thereby broadening the applicability and effectiveness of LeJOT.
\section{Related Work}

\subsection{Cost Optimization}

Cloud cost optimization has been extensively studied, focusing on compute provisioning, storage efficiency, and network cost reduction~\cite{6149074}. Khan et al.~\cite{khan2024cost} proposed a graph-based approach for cloud resource placement, modeling compute and storage instances as nodes and data transfer costs as weighted edges. Using shortest-path algorithms, their method minimizes cost while balancing utilization, performance, and availability.

Recent research highlights data locality-aware orchestration to reduce unnecessary data movement~\cite{s21248212}. Corodescu et al. proposed a container-centric framework, ensuring tasks are executed near their data sources to minimize bandwidth costs. Additionally, AI-driven scheduling techniques predict workload fluctuations, dynamically optimizing resource allocation and data replication~\cite{7835175,10.1145/3136623}.


\subsection{Execution Time Predction}

Research on performance prediction for clusters, grids, or clouds has been an active field for several decades~\cite{295655}. Existing approaches broadly fall into three categories: analytical modeling, simulation/emulation, and empirical evaluation. Recent advancements focus on workflow-specific challenges, leveraging ML and statistical methods to address input variability, cloud heterogeneity, and stochastic execution behaviors.

Miu and Missier~\cite{6495803} showed that input‑aware regression models (MSP, k‑NN, MLP) using features such as dataset size and attribute count reduce C4.5 activity‑time prediction error to~\(\sim20\%\), illustrating both the payoff and the feature‑engineering cost of input predictability.
Chirkin et al.~\cite{CHIRKIN2017376} modeled task runtimes as stochastic variables, propagated the distributions via characteristic functions to estimate workflow makespan, and supplied these estimates to GAHEFT for tighter deadline scheduling. Graph reductions sped computation, and flood‑simulation tests kept makespan error below~10\%.
Pham et al.~\cite{8013738} devised a two‑stage predictor: (i) infer runtime metrics from inputs and VM type, then (ii) feed them with static features into a Random‑Forest that cuts task‑time error to 1.6\% on AWS/Google Cloud and ports to new clouds with minimal retraining.


\section{Conclusion and Discussion}


This study introduced LeJOT, an intelligent job cost orchestration solution tailored specifically for the Databricks platform. By integrating ML techniques for execution time prediction and advanced solver-based optimization algorithms, LeJOT effectively addresses key limitations of conventional scheduling methods. LeJOT successfully provides dynamic and predictive resource allocation, achieving substantial cost reductions while satisfying performance requirements.

Experimental results on real-world Databricks workloads validate the effectiveness of our approach, demonstrating an average reduction in cloud computing costs exceeding 20\%, significantly outperforming traditional strategies. These findings underscore LeJOT's practical value and scalability, making it highly suitable for enterprises aiming to optimize operational costs and improve resource utilization.

Despite its demonstrated success, there remain areas for further improvement and exploration. Future work should consider enhancing adaptability to varying workloads and refining prediction accuracy to handle more diverse and complex job characteristics. Additionally, exploring the extension of LeJOT to multi-cloud environments and GPU-intensive workloads could broaden its applicability and further enhance its practical impact on cost optimization in large-scale data analytics platforms.

\section*{Acknowledgment}

We thank BigCom's anonymous reviewers for their comments and valuable feedback. We also acknowledge Yanan Li, Mingwei Li, Rui Sun, and Ming Zeng. The research is partially supported by National Key R\&D Program of China under Grant  2021ZD0110400, 
Anhui Provincial Natural Science Foundation under Grant 2208085MF172, Innovation Program for Quantum Science and Technology 2021ZD0302900 and China National Natural Science Foundation with No. 62132018, 62231015, ``Pioneer'' and ``Leading Goose''  R\&D Program of Zhejiang,  2023C01029, and 2023C01143. 

\bibliographystyle{IEEEtran}
\bibliography{refer}
\end{document}